\documentclass[french]{pfia}

\usepackage{xcolor}
\usepackage{etoolbox} 
\usepackage{graphicx}
\usepackage{amsmath}
\usepackage{booktabs}  
\usepackage{array}     

\definecolor{afiablue}{RGB}{61,159,207}
\definecolor{afiared}{RGB}{167,75,68}
\definecolor{afialightblue}{RGB}{158,193,232}

\title{\textbf{Profils des utilisateurs souffrant de troubles du sommeil : Vers un clustering explicable et analyse des variables différentielles}}

\author{Sifeddine Sellami\fup{1, 2}, Juba Agoun \fup{1}, Lamia Yessad \fup{2}, Louenas Bounia \fup{3}\\[6pt]
\fup{1} Université Lumière Lyon 2, Laboratoire ERIC, Lyon, France\\
\fup{2} École Nationale Supérieure d'Informatique (ESI), Alger, Algérie\\
\fup{3} LIPN-UMR CNRS 7030 Université Sorbonne Paris Nord, Villetaneuse, France\\}

\date{}

\begin{document}

\maketitle


\begin{resume}
Les troubles du sommeil ont un impact majeur sur la santé et la qualité de vie des patients, mais leur diagnostic reste complexe en raison de la diversité des symptômes. Aujourd’hui, les avancées technologiques, combinées à l’analyse des données médicales, ouvrent de nouvelles perspectives pour une meilleure compréhension de ces troubles. En particulier, \textbf{l’intelligence artificielle explicable (XAI)} vise à rendre les décisions des modèles d’IA compréhensibles et interprétables par les utilisateurs. Dans cette étude, nous proposons une méthode basée sur le \textit{clustering} afin de regrouper les patients selon différents profils de troubles du sommeil. En intégrant une approche explicable, nous identifions les facteurs clés influençant ces pathologies. Une expérimentation sur des données réelles anonymisées illustre l’efficacité et la pertinence de notre approche.
\end{resume}

\begin{motscles}
Troubles du sommeil, Intelligence artificielle explicable (XAI), Clustering, Tests statistiques, Analyse de données.
\end{motscles}

\begin{abstract}
Sleep disorders have a significant impact on patients’ health and quality of life, but their diagnosis remains challenging due to the diversity of symptoms. Today, technological advances combined with medical data analysis offer new opportunities for a better understanding of these conditions. In particular, \textbf{Explainable Artificial Intelligence (XAI)} aims to make AI-driven decisions understandable and interpretable by users. In this study, we propose a clustering-based method to group patients according to different profiles of sleep disorders. By leveraging an explainable AI approach, we identify the key factors that influence these conditions. A case study using anonymized real-world data demonstrates the effectiveness and relevance of our approach.
\end{abstract}



\section{Introduction}
Les troubles du sommeil, qui ont fait l'objet d'une étude scientifique ces dernières années~\cite{Markun2020}, représentent un enjeu important de santé publique en raison de leurs impacts multidimensionnels. Ces altérations affectent non seulement la qualité de vie des individus, mais également leur santé mentale et physique, ainsi que leur productivité au quotidien. Si une détection précoce et une prise en charge adaptée permettent d'améliorer significativement l'état de santé des patients~\cite{Stranges2012}, leur gestion clinique reste particulièrement complexe en raison de la diversité des manifestations symptomatiques et de l'hétérogénéité des causes sous-jacentes~\cite{xu2022review}. Une proportion importante des cas demeure non diagnostiquée et non traitée. Néanmoins, l'émergence des technologies numériques en santé~\cite{Shajari23} et la disponibilité accrue des données médicales, qu'elles proviennent de dispositifs de suivi du sommeil, de dossiers médicaux ou d'applications mobiles, ouvrent des perspectives prometteuses pour améliorer le diagnostic de ces troubles. Comme dans de nombreux domaines, l'IA s'avère un outil prometteur pour \textbf{détecter} et analyser les troubles du sommeil~\cite{xu2022review}. Notre étude s'appuie sur les données de l'application \textbf{KANOPEE}, dédiée au suivi des patients souffrant de \textbf{troubles du sommeil}.
\smallskip

Notre étude vise à segmenter les utilisateurs en groupes homogènes pour identifier les variables influençant leurs troubles du sommeil. Nous appliquons pour cela l'algorithme \textbf{K-means}, puis caractérisons les clusters via trois méthodes complémentaires : (i) tests statistiques, (ii) métriques de distance, et (iii) analyse XAI des frontières inter-clusters. Dans la suite, nous présentons l'application \textbf{KANOPEE} et les données collectées (section \ref{sec:KANOPEE}). Ensuite, nous détaillons notre méthode basée sur le clustering (section \ref{sec:clustering}). Enfin, nous discutons les résultats du protocole expérimental validant notre méthode (section \ref{sec:resultats}) avant de conclure cet article et d'énoncer des perspectives futures (section \ref{sec:conclusion}).

\section{Données KANOPEE} \label{sec:KANOPEE}
\textbf{KANOPEE} est une application mobile gratuite développée par le laboratoire {\sf SANPSY} de l’Université de Bordeaux, en partenariat avec le CHU de Bordeaux. Cette application accompagne numériquement les patients souffrant de troubles du sommeil à l’aide de deux assistants virtuels, \textit{Louise} et \textit{Jeanne}. Ces agents conversationnels réalisent un suivi personnalisé à travers trois visites programmées (J0–V0, J7–V1, J17–V2)\footnote{J : jour depuis l’inscription, V : visite.}.  Les questionnaires administrés permettent de générer des indicateurs cliniques tels que le \textbf{PHQ-9} (échelle de dépression), l’\textbf{ISI} (échelle d’insomnie), entre autres.  Notre analyse des données vise à identifier les variables influençant la \textbf{qualité du sommeil}, regroupées en trois catégories~:

\begin{enumerate}
    \item \textbf{Données initiales :}  âge, sexe, IMC, localisation (département, région), catégorie socio-professionnelle, niveau d'étude, score \texttt{NOSAS} (risque d'apnée), présence de \texttt{SAOS} (syndrome d’apnée du sommeil) ou \texttt{SJSR} (syndrome des jambes sans repos). 
    \item \textbf{Mesures aux visites (J-0, J-7, J-17) :} scores d’anxiété (\texttt{ANX}), de dépression (\texttt{DEP}), d’insomnie (\texttt{ISI}), de somnolence (\texttt{ESS}) et les dates des visites.
    \item \textbf{Variables dérivées :} moyennes et écarts-types des paramètres de sommeil (durée, heures coucher/réveil) pour J-0 : J-7 et J-7 : J-17.
\end{enumerate}

Un problème majeur réside dans la présence de valeurs manquantes, car la plupart des utilisateurs ne complètent pas les trois visites. Afin de garantir la fiabilité des analyses dans un contexte médical, seules les données complètes ont été conservées, aboutissant à un ensemble de \textbf{145 utilisateurs} et \textbf{37 variables}. Dans la suite, les approches $(i), (ii)$ et $(iii)$ ont été appliquées pour détecter les variables influençant le sommeil.

\section{Clustering pour la segmentation des utilisateurs} \label{sec:clustering}

Nous utilisons l’algorithme \textit{K-means} pour regrouper les utilisateurs selon leurs profils de sommeil. Le nombre optimal de clusters $K$ est déterminé à l’aide du meilleur score de silhouette\footnote{Par manque d’espace, nous ne présentons pas le graphique de silhouette ni les scores associés, mais uniquement la valeur optimale de $K$.}, et confirmé visuellement par une analyse en composantes principales (ACP).
\smallskip

Trois méthodes complémentaires permettent de caractériser les clusters et d’identifier les variables influentes :
\smallskip

\begin{itemize}
    \item Des tests statistiques comparatifs entre clusters \cite{kim2018symptom}~
    \smallskip
    \item Une analyse basée sur les distances (\textit{in-pattern/out-pattern}) \cite{khoie2017hospital}~
    \smallskip
    \item Une approche d’intelligence artificielle explicable (XAI) pour interpréter les frontières entre clusters, adaptée de \cite{bobek2022enhancing}.
\end{itemize}
\smallskip

\begin{figure*}[h]
    \centering
    \includegraphics[width=0.95\textwidth]{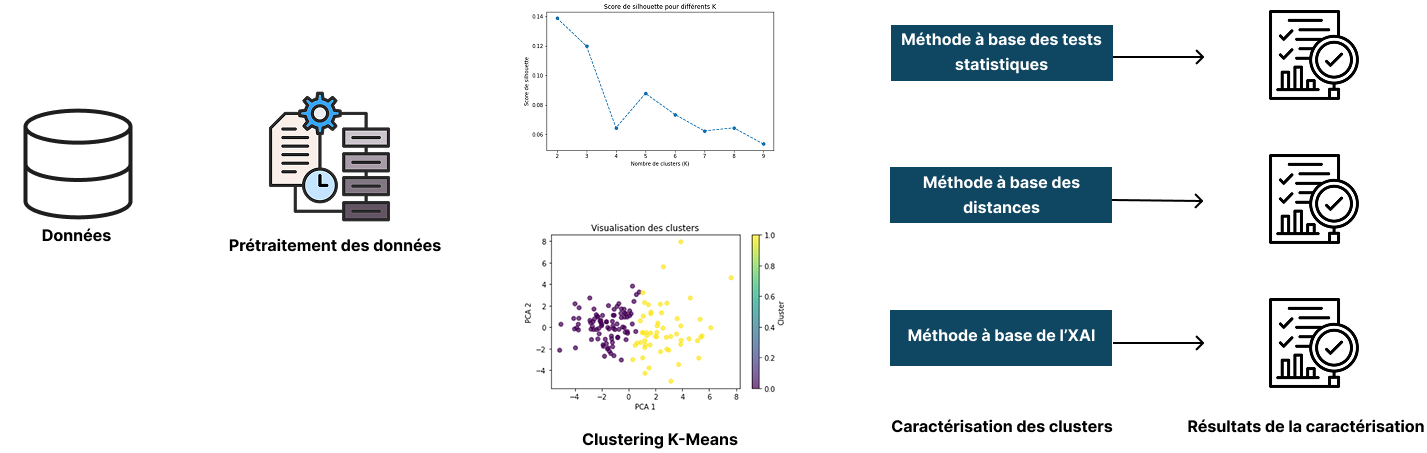}
    \caption{Schéma illustratif de la méthode de segmentation des utilisateurs par clustering}
    \label{fig:meth02}
\end{figure*}

La Figure~\ref{fig:meth02} synthétise notre approche méthodologique, où les méthodes appliquées sont décrites dans les sous-sections ci-après.

\paragraph{Clustering \texorpdfstring{$K$}{K}-means}. Avant l'application de l'algorithme $K$-means \footnote{Le nombre de clusters optimale dans notre cas d’étude est $K = 2$}, un prétraitement a été réalisé incluant l'encodage des variables catégorielles et la standardisation. Le clustering a ensuite permis de constituer des groupes distincts. Après avoir construit les clusters, trois méthodes sont appliquées pour leur caractérisation, c'est-à-dire l'identification des variables saillantes propres à un groupe.

\subsection{Tests statistiques comparatifs} \label{sec:tests_stat}
Cette méthode s’inspire des travaux de \cite{kim2018symptom}. Pour les variables numériques, la moyenne de chaque variable dans chaque cluster est calculée, puis des tests statistiques sont appliqués afin d'évaluer s’il existe une différence significative entre ces moyennes. Le test de {\bf Student} est utilisé lorsque le rapport $\frac{\text{var}_2}{\text{var}_1}$ est compris entre $0{,}5$ et $2$ (hypothèse de variances égales), sinon le test de {\bf Welch} est appliqué. Seules les variables avec une $p$-value $\leq 0{,}05$ sont retenues comme caractéristiques, indiquant une différence significative entre les clusters. Pour les variables catégorielles, nous comparons les pourcentages par catégorie dans chaque cluster à l’aide du test de {\bf Fisher} ou {\bf khi-deux}, selon le même critère. Seules les variables dont la $p$-value $\leq 0{,}05$ sont également retenues.
\smallskip

Une fois les variables caractéristiques identifiées \footnote{Variables présentant des différences significatives entre clusters}, nous interprétons leurs valeurs moyennes cluster par cluster. Pour les indicateurs cliniques \texttt{(ANX, DEP, ISI, ESS, IMC)} dont les échelles standardisées permettent une interprétation absolue, nous analysons les moyennes en référence à ces normes. Pour les autres variables, nous procédons à une comparaison relative entre clusters. Cette analyse détaillée conduit finalement à une caractérisation globale de chaque groupe.

\subsection{Méthode \textit{in-pattern/out-pattern}} \label{sec:distances}

La méthode de \cite{khoie2017hospital} répartit les observations en deux groupes : \textit{in-pattern} et \textit{out-pattern}. Leur analyse permet d'identifier les variables distinctives et leur tendance (valeurs élevées/faibles) pour chaque cluster. Les étapes sont :

\begin{enumerate}
    \item \textbf{Calcul du centroïde} : Détermination de $X_k$ comme moyenne des points du cluster.

    \item \textbf{Classification in/out-pattern} : Pour chaque point $i$, calcul de sa distance euclidienne $d_i$ au centroïde $X_k$. Si $d_i \in [\mu_k \pm z\sigma_k]$, il est \textit{in-pattern}, sinon \textit{out-pattern}, où $\mu_k$ et $\sigma_k$ sont la moyenne et l'écart-type du cluster $k$, et $z$ une constante de seuil.

    \item \textbf{Moyennes in/out-pattern} : Calcul de $\mu_{\text{in}}(k,v)$ et $\mu_{\text{out}}(k,v)$ pour chaque cluster $k$ et variable $v$.

    \item \textbf{Calcul du facteur de différentiation} : Cette étape consiste à calculer, pour chaque variable $v$ et chaque cluster $k$, un facteur de différentiation :
    $$ df(k, v) = \frac{\mu_{\text{in}}(k, v) - \mu_{\text{out}}(k, v)}{\mu_{\text{out}}(k, v)} $$

     Ce facteur mesure la différence relative entre les moyennes des \textit{in-pattern} et \textit{out-pattern}, et permet d’identifier les variables les plus distinctives.

     \item \textbf{Statistiques descriptives} : Calcul des moyennes ($\mu$) et écarts-types ($\sigma$) par cluster, caractérisant la distribution de chaque variable $v$ dans le cluster.

     \item \textbf{Identification des variables saillantes} : Une variable $v$ est dite saillante dans un cluster $k$ si son facteur de différentiation vérifie :
\[
\begin{aligned}
df(k, v) &\leq \mu_{df}(k) - z \sigma_{df}(k) \\
\text{ou} \quad df(k, v) &\geq \mu_{df}(k) + z \sigma_{df}(k)
\end{aligned}
\]

Un facteur positif indique que $v$ prend des valeurs majoritairement élevées dans le cluster $k$, tandis qu’un facteur négatif signifie que ses valeurs y sont majoritairement faibles.
\end{enumerate}

Cette méthode identifie les variables caractéristiques de chaque cluster en indiquant si leurs valeurs y sont majoritairement élevées ou faibles. Ces résultats permettent ensuite de bien caractériser chaque cluster.

\begin{figure*}[ht!]
    \centering
    \includegraphics[width=.99\textwidth]{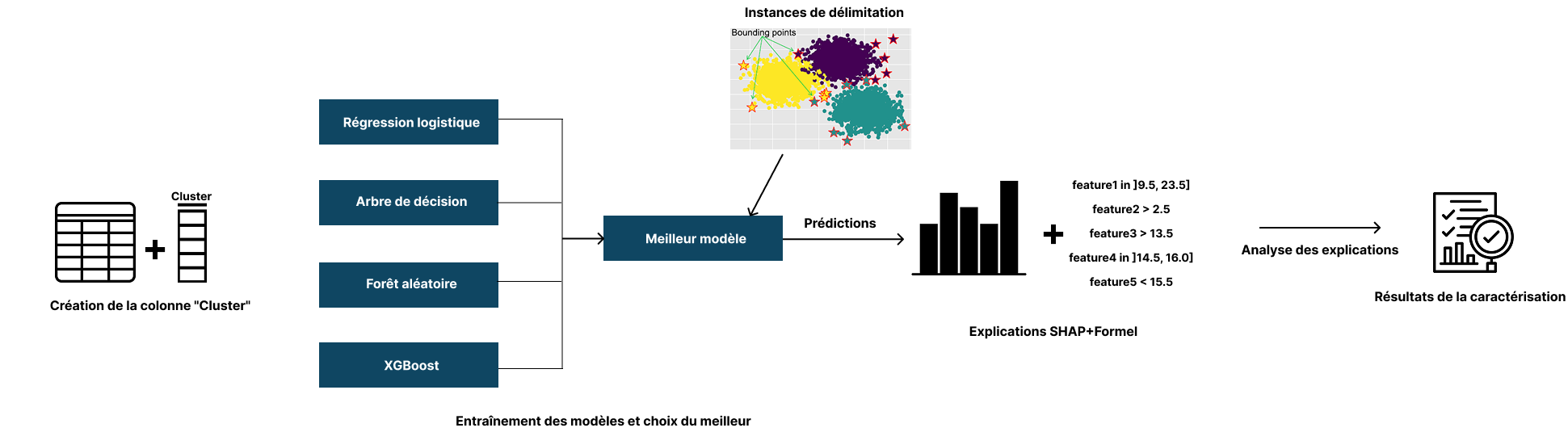}
    \caption{Schéma méthodologique pour l’extraction des variables caractéristiques (approche basée sur l’XAI)}
    \label{fig:meth2.3}
\end{figure*}

\begin{figure*}[ht!]
    \centering
    \includegraphics[width=0.65\textwidth]{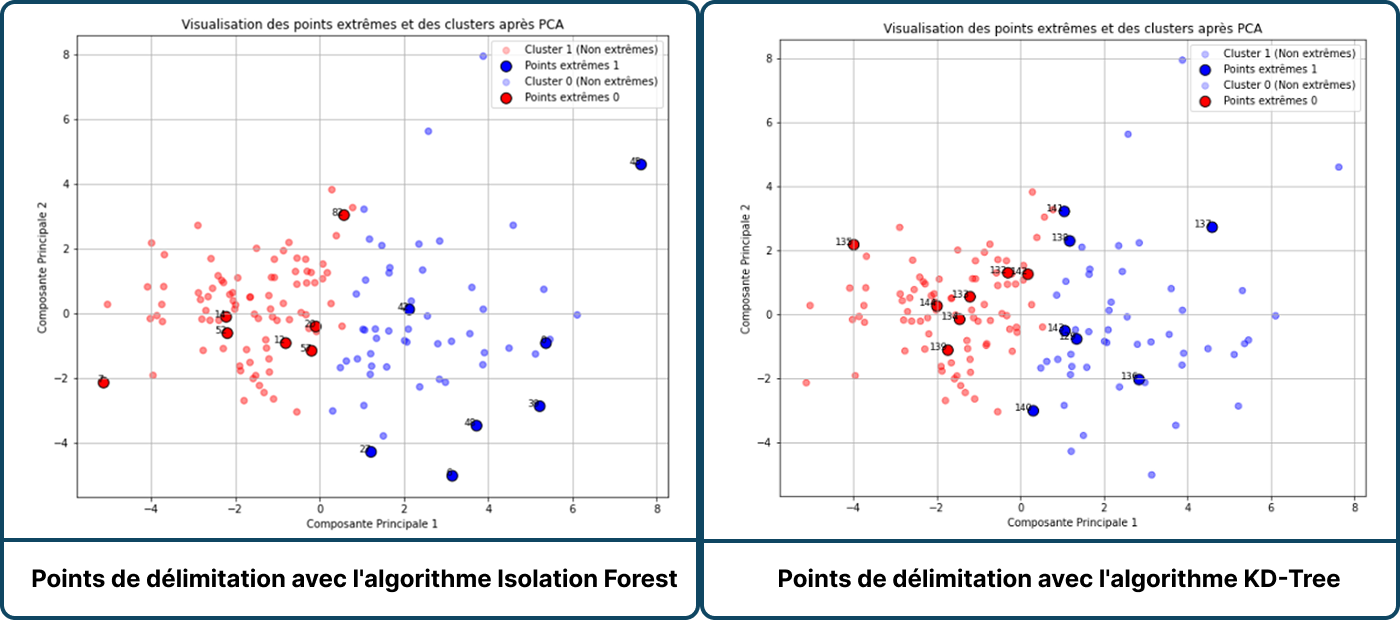}
    \caption{Points de délimitation des clusters identifiés à l'aide des algorithmes KD-Tree et Isolation Forest}
    \label{fig:delimitation}
\end{figure*}

\subsection{Méthode à base de XAI} \label{sec:xai}
S'appuyant sur les travaux de \cite{bobek2022enhancing}, cette méthode applique des techniques de XAI pour analyser et interpréter les clusters. Le schéma global, présenté dans la Figure~\ref{fig:meth2.3}, s'articule autour des quatre étapes suivantes :
\smallskip

\textbf{(a)} Nous appliquons un algorithme de clustering et associons à chaque instance une étiquette correspondant au cluster auquel elle appartient (la variable cible est notée \texttt{Cluster}). \textbf{(b).} Nous entraînons par la suite des classifieurs ($75\%$ de données labellisées pour l'entraînement et $25\%$ pour le test). Pour obtenir des résultats fiables et robustes, nous avons utilisé $4$ modèles : {\bf Régression logistique}, {\bf Arbre de décision}, {\bf Forêt aléatoire} et {\bf XGBoost}. Ensuite, nous avons sélectionné pour chaque cluster le modèle le plus performant en termes de précision et de score $F_1$. \textbf{(c).} Avant de dériver les variables influentes, nous nous intéressons aux points de délimitation. En fait, selon \cite{bobek2022enhancing}, les meilleurs points à expliquer sont les points de délimitation dans chaque cluster. C'est pourquoi nous avons appliqué la méthode \textit{KD-Tree} et \textit{Isolation Forest} pour déterminer ces points de délimitation (une visualisation de ces points a été faite à l'aide d'une \textit{ACP} via la figure \ref{fig:delimitation}). Nous avons sélectionné uniquement $5\%$ des instances comme points de délimitation. \textbf{(d)}. Après avoir défini les instances à expliquer, nous avons opté pour une méthode qui combine \textbf{SHAP} \cite{Lundberg17} et les explications formelles générées par l'outil (PyXAI) \cite{AudemardPyXAI} \footnote{Voir \url{https://www.cril.univ-artois.fr/pyxai/} pour plus de détails sur {\bf PyXAI} et les explications formelles}. Notre choix est bien détaillé et justifié dans la section~\ref{sec:SHAPFormel} ({\sf SHAP + Formel)}.
\smallskip

\textbf{En résumé}, notre approche consiste d’abord à identifier les variables clés influençant la formation des clusters, puis à les interpréter selon leur nature : à l’aide de seuils cliniques pour les scores médicaux (\textbf{ANX}, \textbf{DEP}, etc.), ou par comparaison aux moyennes pour les autres variables. L’analyse des cas frontières permet enfin de dégager des profils types représentatifs de chaque groupe.

\subsection{Méthode SHAP + Formel}\label{sec:SHAPFormel} 

Dans cette sous-section, nous proposons l’approche \textbf{SHAP + Formel}, qui combine la méthode SHAP avec des explications formelles de type abductif (voir \cite{JoaoLogicXAI, AudemardKR21} pour un état de l’art détaillé des méthodes formelles en XAI). La méthode SHAP \cite{Lundberg17} est une technique d’attribution de caractéristiques qui quantifie la contribution de chaque variable à une prédiction donnée. Cependant, cette approche présente deux limites majeures : $(1)$ elle ne prend pas en compte les interactions entre les variables, supposant leur indépendance, une hypothèse rarement vérifiée en pratique, notamment dans notre contexte, $(2)$ sa fiabilité demeure limitée dans les domaines critiques tels que la médecine \cite{JoaoLogicXAI, dke, Audemard22ijcai}. Par ailleurs, les scores d’importance générés par SHAP ne permettent pas de distinguer clairement les clusters identifiés.
\smallskip

D’un autre côté, les explications formelles, réputées plus fiables car fondées sur la logique mathématique \cite{JoaoLogicXAI}, sont généralement exprimées sous forme de règles. Toutefois, même lorsqu’elles sont minimales, ces règles peuvent être relativement longues (souvent composées de $7 \pm 2$ éléments), ce qui en complique l’interprétation \cite{Miller56}, en particulier pour la caractérisation des clusters. L’idée est donc de combiner les explications fournies par {\bf SHAP}, qui mettent en évidence les variables les plus influentes, avec les explications formelles. Concrètement, nous calculons l’intersection entre les $10$ variables les plus importantes selon SHAP et celles présentes dans les explications formelles, afin d’obtenir une règle à la fois courte, fiable, correcte, composée de variables influentes et facilement interprétable.

\section{Expérimentations et Résultats} \label{sec:resultats}
Cette section présente les résultats du clustering K-means (optimal pour $k=2$, confirmé par le score de silhouette et la visualisation ACP), visant à identifier les facteurs des troubles du sommeil. Après prétraitement des données, trois méthodes complémentaires ont caractérisé les clusters et leurs variables les plus discriminantes.

\subsection{Caractérisation par tests statistiques} 

Après application des tests statistiques décrits dans la section \ref{sec:tests_stat} pour identifier les variables présentant des différences significatives entre les deux clusters, aucune variable catégorielle ne s’est révélée statistiquement significative. En revanche, \textbf{certaines variables numériques ont montré des différences significatives} entre les deux groupes. Les variables avec les différences les plus significatives sont répertoriées dans le Tableau~\ref{tab:tests_stat}.

\begin{table}[h]
    \centering
    \small
    \begin{tabular}{lccc}
    \toprule
    \textbf{Variable} & \textbf{Cluster 0} & \textbf{Cluster 1} & \textbf{Différence} \\
    \midrule
    Age & $52.04 \pm 12.13$ & $47.14 \pm 12.95$ & $-4.90$ \\
    ESSV0 & $6.68 \pm 4.27$ & $10.25 \pm 5.03$ & $3.40$ \\
    ISIV0 & $13.12 \pm 4.32$ & $17.44 \pm 3.68$ & $4.31$ \\
    DEPV0 & $6.07 \pm 3.37$ & $13.25 \pm 4.24$ & $7.19$ \\
    ANXV0 & $2.52 \pm 1.63$ & $4.51 \pm 1.35$ & $1.99$ \\
    TSTISDV0 & $2.30 \pm 0.92$ & $2.87 \pm 1.03$ & $0.58$ \\
    MIDISDV0 & $1.53 \pm 0.72$ & $1.99 \pm 1.28$ & $0.46$ \\
    SRIV0 & $85.59 \pm 5.94$ & $81.98 \pm 7.73$ & $-3.61$ \\
    \bottomrule
    \end{tabular}
    \caption{Comparaison inter-clusters des variables significatives (moyenne $\pm$ écart-type). Les différences positives indiquent des valeurs plus élevées dans le Cluster $1$.}
    \label{tab:tests_stat}
\end{table}
\smallskip

L'analyse du tableau \ref{tab:tests_stat} selon les seuils cliniques révèle des différences marquées entre clusters. Les patients du Cluster $1$ sont plus jeunes (AGE) et présentent : une somnolence diurne excessive (ESS), une insomnie modérée (ISI), une dépression modérée (DEP) avec risque d'épisode dépressif majeur (ANX), ainsi qu'un sommeil irrégulier (faible SRIV, forte dispersion TSTISD et MIDISD).
\smallskip

À l'inverse, le Cluster $0$ regroupe des individus plus âgés (AGE) avec : une somnolence normale (ESS), une insomnie légère subclinique (ISI), une dépression légère (DEP) sans anxiété significative (ANX), et un sommeil régulier (SRIV élevé et faible dispersion des paramètres de sommeil).

\paragraph{Bilan.}~ Les résultats révèlent une dichotomie clinique nette entre les deux clusters. Le Cluster $1$ se caractérise par une population jeune présentant une symptomatologie marquée associant dépression clinique, insomnie modérée et altération de la régularité du sommeil, tandis que le Cluster $0$ regroupe des sujets plus âgés avec des troubles de l'humeur atténués et un sommeil préservé tant en qualité qu'en régularité, reflétant ainsi un profil clinique plus favorable.

\subsection{Caractérisation à base des distances}
En appliquant la méthode décrite dans la section \ref{sec:distances}, fondée sur le calcul des in-patterns et out-patterns, nous avons reproduit l'approche de \cite{agoun2023opticlust4rec} qui nous a permis d'identifié les variables caractéristiques de chaque cluster selon leurs valeurs généralement élevées ou faibles.

Pour le \textbf{Cluster~0}, les variables à valeurs majoritairement élevées sont l’âge, \textit{SRIV0} et \textit{SRIV1}, indiquant une population plus âgée avec un sommeil plus régulier. En revanche, les valeurs majoritairement basses concernent \textit{ISIV1}, \textit{ISIV2}, \textit{DEPV0}, \textit{DEPV1}, \textit{DEPV2}, \textit{ANXV1} et \textit{ANXV2}, suggérant de faibles niveaux d’insomnie, de dépression et d’anxiété.
\smallskip

Pour le \textbf{Cluster~1}, les valeurs élevées portent sur \textit{ISIV1}, \textit{ISIV2}, \textit{DEPV0}, \textit{DEPV1}, \textit{DEPV2}, \textit{ANXV1}, \textit{ANXV2} et \textit{CRP}, traduisant une forte prévalence d’insomnie, de dépression, d’anxiété et une possible inflammation. À l’inverse, l’âge (\textit{Age}) y est plus faible, indiquant une population plus jeune.
\smallskip

\paragraph{Bilan.}~ L’analyse révèle deux profils distincts : le Cluster $1$ regroupe des patients jeunes avec une triade symptomatique (dépression clinique, troubles insomniaques, activité professionnelle), suggérant un lien possible avec le stress professionnel. À l’inverse, le Cluster $0$ correspond à des sujets âgés présentant une dépression atténuée et un sommeil régulier, reflétant un équilibre mental.

\subsection{Caractérisation à base de XAI}
Après avoir réalisé le clustering et ajouté la colonne \textit{Cluster} au jeu de données (cf. section~\ref{sec:xai}), nous avons entraîné quatre modèles de classification dont les performances, optimisées via {\bf Optuna}\footnote{Optuna est un framework d'optimisation d'hyperparamètres}, sont présentées dans le tableau~\ref{tab:models_clf}. La forêt aléatoire s'étant révélée la plus performante, nous l'avons sélectionnée pour les étapes ultérieures d'explicabilité. Afin de caractériser les clusters, nous avons déterminé leurs frontières à l'aide des algorithmes KD-Tree \footnote{Le kd-Tree est une structure permettant d'organiser des données présentes dans un espace a $k$-dimensions selon leur répartition spatiale.} et Isolation Forest \footnote{L'isolation forest calcule un score d'anomalie pour chaque instance.} en sélectionnant $5\%$ du jeu de données (soit $7$ instances par cluster) comme points de délimitation, dont la répartition est visualisée dans la figure~\ref{fig:delimitation}.

\begin{table}[h]
    \centering
    \small
    \begin{tabular}{lccc}
        \toprule
        \textbf{Modèles} & \textbf{Accuracy} & \textbf{F\textsubscript{1} (C\textsubscript{0})} & \textbf{F\textsubscript{1} (C\textsubscript{1})} \\
        \midrule
        Régression logistique & 97.29\% & 98\% & 96\% \\
        Arbre de décision & 83.78\% & 88\% & 75\% \\
        Forêt aléatoire & \textbf{97.30\%} & \textbf{98\%} & \textbf{96\%} \\
        XGBoost & 94.59\% & 96\% & 92\% \\
        \bottomrule
    \end{tabular}
    \caption{Performances comparées des modèles de classification sur les clusters C\textsubscript{0} et C\textsubscript{1}.}
    \label{tab:models_clf}
\end{table}

Après avoir identifié ces points, les explications {\bf SHAP+Formel} (voir section \ref{sec:SHAPFormel}) ont révélé les variables clés influençant la prédiction pour chaque cluster. Leur interprétation s'appuie sur les intervalles médicaux standards \cite{xu2022review}.
\smallskip

\paragraph{Cluster $0$ (Instance $142$).~} Les variables clés pour cette prédiction sont : DEPV0, DEPV1, DEPV2, ANXV0, ANXV1, ANXV2, ISIV0, ISIV1, TSTISDV0, AGE. Cette personne présente une dépression minimale (DEP), aucune anxiété (ANX), une insomnie sub-clinique légère (ISI), une faible dispersion du temps de sommeil (TSTISDV0) et un âge supérieur à la moyenne.
\smallskip

\paragraph{Instance $143$ (Cluster $1$).~} Les variables clés pour cette prédiction sont : DEPV0, DEPV1, ISIV1, ANXV1, TSTISDV1, ESSV0, ESSV1, AGE. Cette personne présente une dépression modérée (DEP), un probable épisode dépressif majeur (ANX), une somnolence diurne excessive marquée (ESS), une insomnie clinique modérée (ISI), une forte dispersion du temps de sommeil (TSTISDV1), et un âge inférieur à la moyenne.
\smallskip

\paragraph{Bilan.}~ L'analyse des interprétations des instances à la frontière entre les deux clusters montre que les individus du cluster $0$ sont généralement plus âgés, moins déprimés et présentent un sommeil régulier. À l'inverse, ceux du cluster $1$ sont plus jeunes, souffrent de dépression et d’insomnie, et ont un sommeil irrégulier et insuffisant.

\section{Conclusion} \label{sec:conclusion}
Les trois méthodes de caractérisation des clusters donnent des résultats très similaires. Il en ressort que nos données identifient deux groupes aux caractéristiques distinctes :
\smallskip

\begin{itemize}
\item Un groupe de jeunes individus déprimés, souffrant d’insomnie, sommeil irrégulier et insuffisant.
\smallskip
\item Un groupe de personnes âgées, moins déprimées, ayant un sommeil régulier et suffisant.
\end{itemize}
\smallskip

Cependant, les approches utilisées pour obtenir ces résultats, ainsi que la forme des résultats, diffèrent selon la méthode. Les deux premières permettent uniquement d’identifier les variables saillantes de chaque cluster. En revanche, la méthode basée sur XAI se distingue par sa capacité à fournir des informations plus riches et interprétables. Elle permet aux utilisateurs de visualiser la contribution de chaque variable (via SHAP) et de voir la règle caractérisant l’état de l’utilisateur, notamment pour les instances de délimitation, facilitant ainsi la compréhension de l’influence des variables sur l’appartenance à un cluster. Cela renforce aussi la compréhension et la confiance des utilisateurs dans l’interprétation des clusters.
\smallskip

\paragraph{Perspective.}~ Une ouverture prometteuse serait d'améliorer la gestion des données manquantes en enrichissant l'ensemble de données. L'approche XAI proposée pourrait être approfondie pour reconstruire les valeurs manquantes avec les informations disponibles, tout en maintenant l'interprétabilité des résultats.

\section*{Remerciements}
Nous remercions sincèrement les rapporteurs pour leurs commentaires et suggestions. Ce travail a bénéficié d’une aide de l’État gérée par l’Agence nationale de la recherche au titre de France $2030$ portant la référence  {\bf « ANR-21-SFRI-0001 »}.

\bibliographystyle{plain}

\end{document}